%% file: main.tex
\documentclass[journal]{IEEEtran} %[journal] instead of draft

\usepackage[utf8]{inputenc}         % allow utf-8 input
\usepackage[T1]{fontenc}            % use 8-bit T1 fonts
\usepackage{hyperref}               % hyperlinks
\usepackage{url}                    % simple URL typesetting (links)
\usepackage{nicefrac}               % compact symbols for 1/2, etc.
\usepackage{microtype}              % microtypography
\usepackage{amsfonts}               % blackboard math symbols
\usepackage{amsmath}                % For math
\usepackage[dvipsnames]{xcolor}     % For colored text
\usepackage{float}                  % For using H in images
\usepackage{graphicx}               % Specifies path and format of images
\graphicspath{{images/}}
\DeclareGraphicsExtensions{.pdf,.PDF,.jpg,.JPG,.jpeg,.JPEG,.png,.PNG}
\usepackage{booktabs}               % professional-quality tables
\usepackage{multicol}               % Figure on multicolumns 
\usepackage{multirow}               % Figure on multirow
\usepackage{booktabs}               % Fancy tables
\usepackage{array}                  % Fancy tables
\newcolumntype{C}[1]{>{\centering\arraybackslash}p{#1}}

\usepackage[nolist]{acronym}        % Abbreviations
\usepackage{verbatim}               % Comments
\usepackage{comment}                % Comment environment

\usepackage{textcomp}
\usepackage[numbers]{natbib}
\usepackage{makecell}
\usepackage[super]{nth} % 1st, 2nd, 3rd etc.
\usepackage{amssymb}    % Symbols I think
\usepackage{graphicx}   % Symbols again I think
\usepackage{subcaption} % Subfigures

\newcommand{\etal}{\textit{et al}.,}

\newcommand{\eg}{\textit{e}.\textit{g}.}

\usepackage{xcolor}
\newcommand\todo[1]{\textcolor{blue}{TODO: #1}}
\newcommand\tocite[1]{\textcolor{blue}{[REFERENCE]}}

\begin{document}

% Paper title
\title{Deep Perceptual Similarity is Adaptable to Ambiguous Contexts}

\author{
    \IEEEauthorblockN{
        \textbf{Gustav~Grund~Pihlgren}\IEEEauthorrefmark{1}, \and
        \textbf{Fredrik~Sandin}\IEEEauthorrefmark{1},
        \and \textbf{Marcus~Liwicki}\IEEEauthorrefmark{1}
    }\\
    \vspace{0.2cm}

    \IEEEauthorblockA{
        \IEEEauthorrefmark{1}%
        \textit{Machine Learning Group} \\
        Lule{\aa} University of Technology, Sweden\\
        %\{firstname\}.\{lastname\}@ltu.se\\
        %\vspace{0.15cm}
    }
}

% after the title page when using the two side option.
\markboth{}{First author et al. : Title}

% Make the title area
\maketitle

\thispagestyle{empty}

%%*************************************************************************
% Abstract
\input{sections/0_abstract}

% Introduction 
\input{sections/1_introduction.tex}

%% Related Work 
%\input{sections/2_related_work.tex}

% DPS, feature analysis, and Evaluation procedure
\input{sections/3_experimental_setup.tex}

% Results
\input{sections/4_results.tex}

% Discussion and Future Work
\input{sections/5_analysis.tex}

% Appendices 
%\section*{Acknowledgment}
%The work presented in this paper has been partially supported by TODO funded by TODO with the grant number TODO.
%\textit{redacted for blind submission} 

% Bibliography
%\typeout{}
\bibliography{biblio}
\bibliographystyle{IEEEtranN}

\end{document}

%% file: sections/0_abstract.tex
\begin{abstract}
% Typically you would want to have one sentence for the following concepts:
% - The problem tackled
% - Why nobody else has adequately answered the research question yet
% - How you tackled the research question
% - How did you go about doing the research that follows from your big idea
% - What’s the key impact of your research?

The concept of image similarity is ambiguous, and images can be similar in one context and not in another.
This ambiguity motivates the creation of metrics for specific contexts.
This work explores the ability of deep perceptual similarity (DPS) metrics to adapt to a given context.

DPS metrics use the deep features of neural networks for comparing images.
These metrics have been successful on datasets that leverage the average human perception in limited settings.
But the question remains if they could be adapted to specific similarity contexts.

No single metric can suit all similarity contexts, and previous rule-based metrics are labor-intensive to rewrite for new contexts.
On the other hand, DPS metrics use neural networks that might be retrained for each context.
However, retraining networks takes resources and might ruin performance on previous tasks.

This work examines the adaptability of DPS metrics by training ImageNet pretrained CNNs to measure similarity according to given contexts.
Contexts are created by randomly ranking six image distortions.
Distortions later in the ranking are considered more disruptive to similarity when applied to an image for that context.
This also gives insight into whether the pretrained features capture different similarity contexts.
The adapted metrics are evaluated on a perceptual similarity dataset to evaluate if adapting to a ranking affects their prior performance.

The findings show that DPS metrics can be adapted with high performance.
While the adapted metrics have difficulties with the same contexts as baselines, performance is improved in $99\%$ of cases.
Finally, it is shown that the adaption is not significantly detrimental to prior performance on perceptual similarity.

The implementation of this work is available online\footnote{\url{https://github.com/LTU-Machine-Learning/Analysis-of-Deep-Perceptual-Loss-Networks}}.

\end{abstract}

%% file: sections/1_introduction.tex
\section{Introduction}
\label{toc:introduction}

% Basically another abstract, but this time longer and with references.

% The problem tackled
% Why is it a problem on the first place i.e. why did not someone fix it already? In other words: what makes it hard?

% Why would one care about this problem?
% Why are you even making this ? Why does it makes sense? 

% Why are we performing the study?

% Explain how we limit the study.

% In the end, what do you bring new? Why would one care?

The ability to measure the similarity of images is fundamental to many tasks and methods in computer vision.
Similarity is an ambiguous concept, and as such, research on image similarity has focused on so-called perceptual similarity, where the goal is to approximate human (or animal) perception of similarity.
Perceptual similarity metrics can be directly applied to tasks such as image retrieval~\cite{hsu1995integrated} and image quality assessment~\cite{kazmierczak2022study}.

Recently a method called deep perceptual similarity (DPS) has achieved close to human performance on perceptual similarity~\cite{zhang2018unreasonable, kumar2022do}.
DPS metrics compare the difference between the deep features (activations) of a neural network when the input is one image compared to another.
This approach has been used to calculate the loss of machine learning models with image outputs, a practice referred to as deep perceptual loss (DPL).
DPL has been successfully applied to image synthesis and transformation such as image generation~\cite{larsen2016autoencoding}, style-transfer~\cite{gatys2016image}, and super-resolution~\cite{johnson2016perceptual}.
DPL has also been used for tasks with image-like outputs such as image segmentation~\cite{mosinska2018beyond} and depth prediction~\cite{liu2021perceptual}.

The neural network from which the deep features are extracted is referred to as the loss network in DPL applications.
For cohesion, the neural networks used for similarity calculation in DPS metrics will also be called loss networks.

However, perceptual similarity has long been known to be
an ambiguous concept~\cite{smith1992perceptual}, with the perception of similarity varying between populations and even within individuals as the context or focus changes.
This issue has not been addressed as metrics have struggled to keep up with human performance, even on tasks where humans tend to agree and not change their perception.
However, as DPS metrics push performance closer to human-level on these datasets, the issue of ambiguity is ripe for evaluation.
Additionally, with the rise of DPL, the performance of the metrics can be measured not only by adherence to human judgments but also by the downstream performance of the models being trained with the loss.
It has already been shown that there is no strong correlation between loss networks performing well for DPS and them being useful for DPL~\cite{pihlgren2023systematic}.
The many different image domains, the context of the image collection, and the goal of the downstream task influence how similarity should be measured.
For example, in medical imaging, the different techniques of preparing samples may lead to significant differences in color that in other contexts might be indicative of dissimilarity but likely has little relevance for diagnosis.
In fact, DPL has already been specifically used to handle this issue of varying staining~\cite{sjostrand2018cell}.

The issue of ambiguity in perceptual similarity raises many exciting research directions, some of which are discussed in Section~\ref{toc:discussion}.
One of those directions regards the ability of different perceptual similarity metrics to adapt to varying definitions of similarity.
Some rule-based metrics are unchanging and therefore unable to adapt~\cite{eskicioglu1995image}.
Other rule-based metrics have hyperparameters that can be altered to fit the metric to particular circumstances, though the hyperparameters are typically limited in how they can alter the metrics~\cite{eskicioglu1995image}.
DPS metrics, on the other hand, are based on neural networks that could theoretically be retrained to suit the particular circumstance the metric is used in.

A problem with training a neural network for each circumstance which DPS and DPL are used for is that this would be resource intensive.
DPS is commonly implemented with pretrained networks, and the most common uses of DPL utilize pretrained networks as well.
These networks are typically pretrained using ImageNet~\cite{deng2009imagenet}, an image dataset the size of which makes it computationally intensive to train on.
However, it is possible that no retraining is needed.
The deep neural networks used in DPS and DPL learn a large number of features that might be useful for many different definitions of similarity.
So rather than retraining the network itself, a layer of scalars could be learned to balance the relevant features for a given circumstance.
\textit{Zhang} \etal{}~\cite{zhang2018unreasonable} showed that this could be used to improve the performance of perceptual similarity for the specific image-distortion distribution that the networks were trained on, but that the improvements did not generalize to other distributions.
Though, ImageNet pretrained CNNs have been shown to be biased toward the texture of the image over other structures~\cite{geirhos2019imagenet}.
Such bias could potentially be problematic for tuning pretrained networks to contexts where textures are less important.

In addition to concerns regarding retraining, there is a possibility that the convolutional neural network (CNN) architectures that are typically used in DPS and DPL cannot be adapted to certain circumstances.
CNN architectures are known to have flaws that make them vulnerable to certain distortions of the input~\cite{azulay2019why}.
Since these flaws are related to the architecture, training a specific model to overcome them might not be possible.

This work investigates whether DPS metrics can be adapted to specific definitions of similarity by training scalars to learn which of the extracted features are relevant for a given definition.
To do this, six common image distortions are used to generate different similarity contexts.
Each context is created by randomly ranking the six distortion types and then defining an image distorted by one distortion to be more similar to the original than if it had been distorted by another later in the ranking.
The metrics that are evaluated in this work consist of the combinations of three pretrained CNN architectures with defined feature extraction layers and five different methods for comparing the extracted features. 
The metrics are then adapted to each definition by training the scalars in the same way as was done by \textit{Zhang} \etal{}~\cite{zhang2018unreasonable}, to recognize some distortions as more similar than others.
The images that are distorted for training are taken from the Street View House Numbers (SVHN) dataset~\cite{netzer2011reading}.
The adapted metrics are then evaluated on how well they recognize the correct distortions as more similar using images from the test sets of SVHN and STL-10~\cite{coates2011analysis} datasets.
Additionally, the adapted metrics are evaluated on the Berkeley-Adobe Perceptual Patch Similarity (BAPPS) dataset~\cite{zhang2018unreasonable} to see how the adaption affects their performance on known human judgments.
In addition to the adapted metrics, the baseline metrics without adaption are also evaluated for reference.

The results of this evaluation show that the metrics could adapt to the different contexts and outperform baseline metrics. 
The performance of baseline and adapted metrics on the same ranking are shown to be correlated, meaning they perform well on the same rankings.
Additionally, the adaption training has only a slim detrimental effect on the perceptual similarity performance on BAPPS.

Potential improvements to the method are discussed, such as allowing negative scalars to handle the rankings that baseline metrics struggle with.
Finally, a broader discussion about ambiguity in perceptual similarity and the potential applications and relevance of adaptability.
For example, comparisons are made to the field of contrastive learning in which different training methods adapted to specific domains are being explored~\cite{chandra2023self}.
Taking inspiration from contrastive learning, an improvement to training metrics is proposed. 

%% file: sections/3_experimental_setup.tex
%%%%%%%%%%%%%%%%%%%%%%%%%%%%%%%%%%%%%%%%%
% Datasets section
%%%%%%%%%%%%%%%%%%%%%%%%%%%%%%%%%%%%%%%%%

\section{Datasets}
\label{toc:datasets}

Three datasets are used in this work, SVHN~\cite{netzer2011reading}, STL-10~\cite{coates2011analysis}, and BAPPS~\cite{zhang2018unreasonable}.

\begin{figure}[t] % Always use [t] or [!t]
    \centering
    \includegraphics[width=\columnwidth]{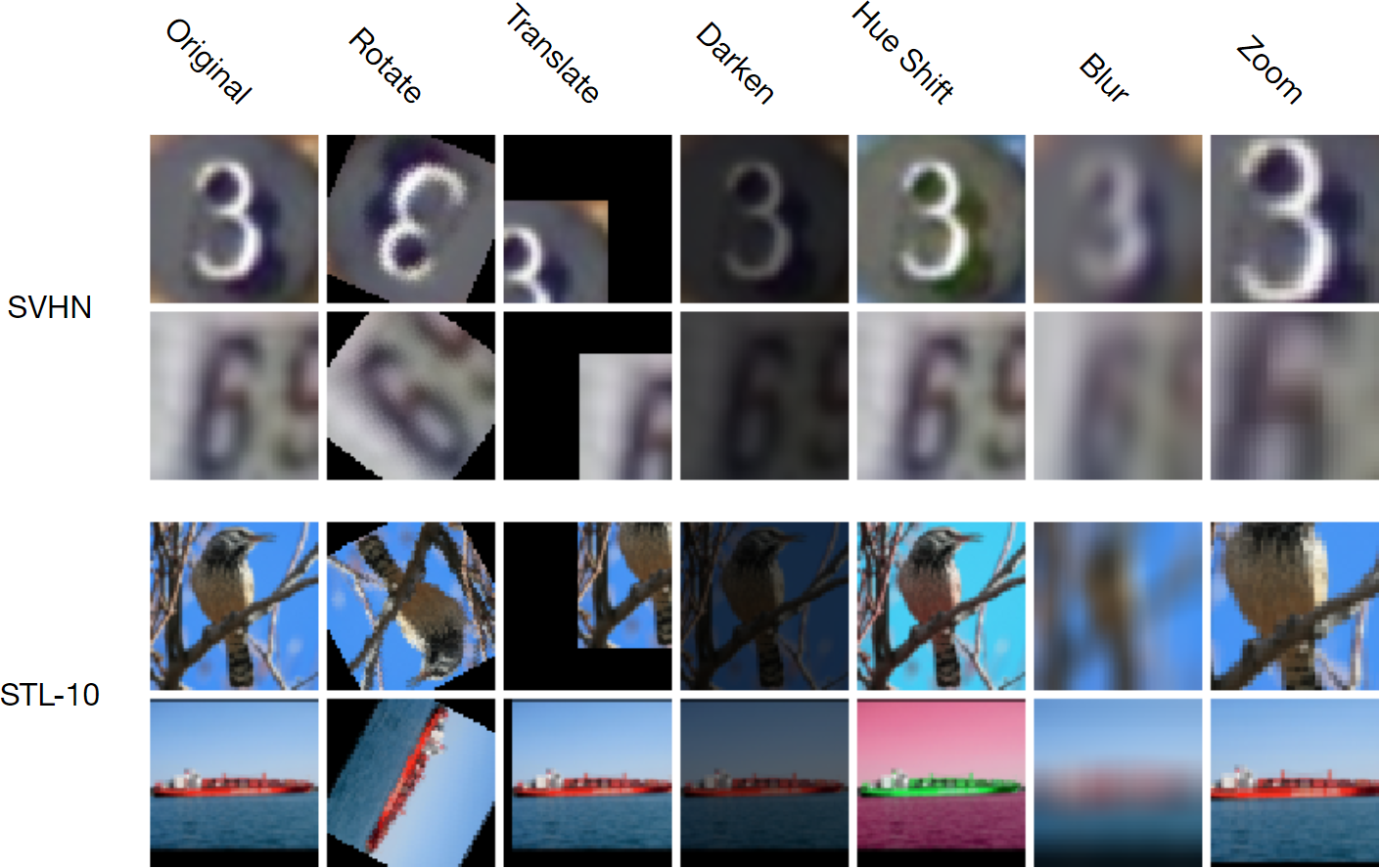}
    \caption{
        Example images from the test sets of SVHN and STL-10 with the six distortion types used in this work applied.
    }
    \label{fig:svhn_and_stl10_distortions}
\end{figure}

SVHN and STL-10 are image classification datasets but are not used for that purpose in this work.
Instead, the SVHN images are used for adaption training to a given ranking of distortions and to test how well the metrics perform.
STL-10 is used only for testing the metrics performance for a given ranking of distortions.
The datasets are used since SVHN and STL-10 consist of significantly different images, which tests whether the adaption learned on one image distribution generalizes to different distributions.
The difference between the datasets can be seen in Fig.~\ref{fig:svhn_and_stl10_distortions}, which shows samples from both, along with applications of the six distortion types that are detailed later.

BAPPS is a perceptual similarity dataset used to evaluate how well a metric estimates human perception of similarity.
%This work uses BAPPS in the same way as it was originally, except no training is done on the dataset.
BAPPS is used in this work to evaluate how the adaption training affects the metrics performance as estimators of human perception of similarity.
As such, it is only used for testing, and no training is done on the dataset.

\subsection{SVHN}
SVHN~\cite{netzer2011reading} consists of photos of house numbers.
The dataset is available with both the original photos and cropped $32\times32$ pixel images of the individual digits.
This work uses cropped images, which include $73257$ digits for training and $26032$ for testing.
The training images, without their labels, are used for adaption training of the metrics.
The testing images are used to evaluate how well the various metrics can decide which distortions are more similar, as defined by a random ranking of the distortions.

\subsection{STL-10}
STL-10~\cite{coates2011analysis} consists of photos of animals and vehicles taken from the ImageNet~\cite{deng2009imagenet} dataset and scaled down to $96\times96$ pixels.
This work uses the $8000$ testing images, without their labels, to evaluate how well the metrics have adapted to the random ranking of distortions they were trained on.

\subsection{BAPPS}

\begin{figure}[t] % Always use [t] or [!t]
    \centering
    \begin{subfigure}[b]{0.5\columnwidth}
        \centering
        \includegraphics[width=\columnwidth]{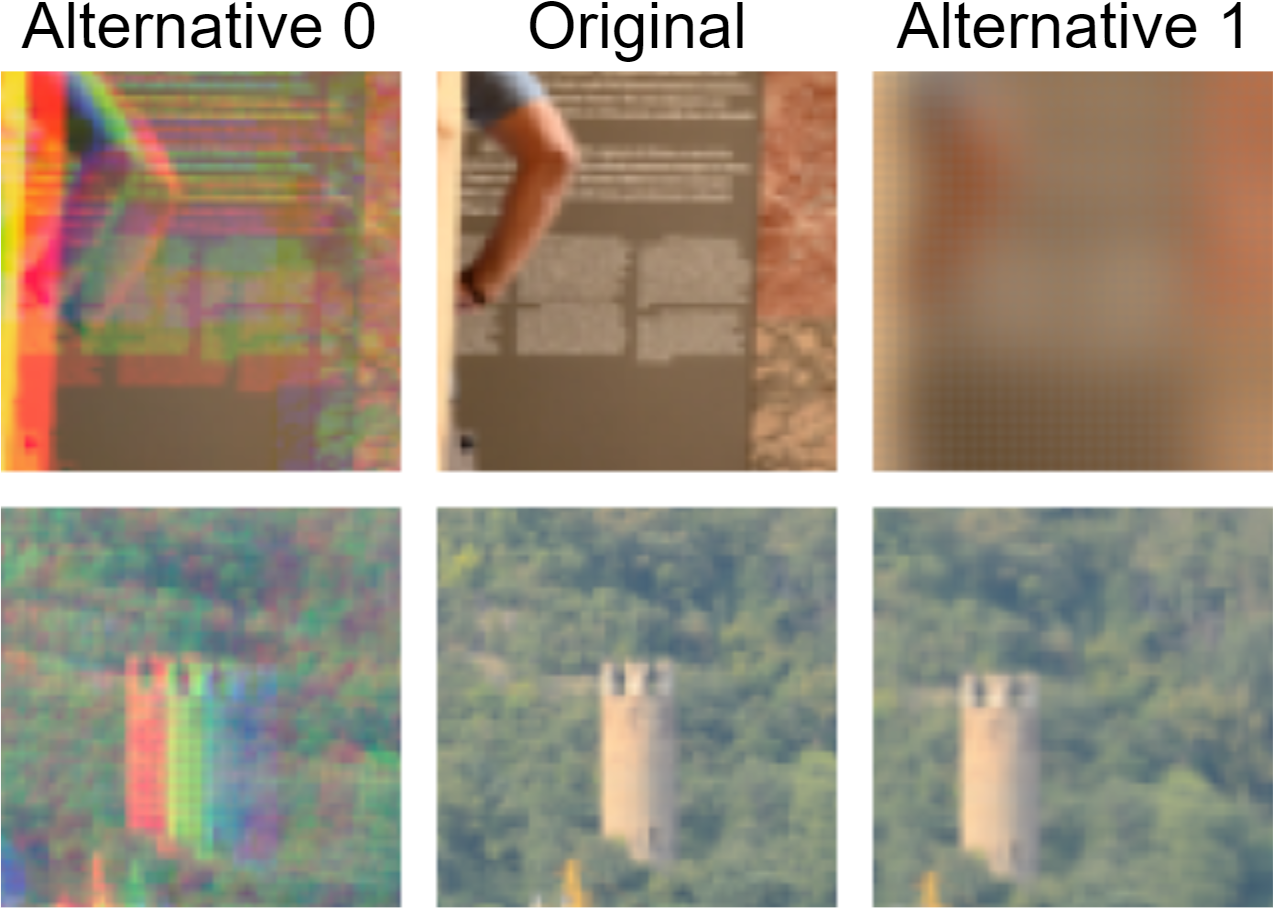}
        \caption{}
    \end{subfigure}
    \hspace{16pt}
    \begin{subfigure}[b]{0.33\columnwidth}
        \centering
        \includegraphics[width=\columnwidth]{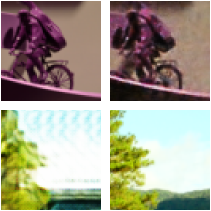}
        \caption{}
    \end{subfigure}
    \caption{
       Sample data from the (a) 2AFC and (b) JND portions of the BAPPS dataset.
       The task on the 2AFC part is to identify which alternative humans perceive as more similar to the original.
       The task on the JND part is to give similarity scores to each pair of images that rank all pairs according to the fraction of humans who mistook the images as being the same.
    }
    \label{fig:bapps_samples}
\end{figure}

BAPPS consists of $64\times64$ image patches sampled from the MIT-Adobe 5k~\cite{bychkovsky2011learning}, RAISE1k~\cite{dang2015raise}, DIV2K~\cite{agustsson2017ntire}, Davis Middleburry~\cite{scharstein2001taxonomy}, video deblurring~\cite{su2017deep}, and ImageNet~\cite{deng2009imagenet} datasets.
The various image patches have been distorted using methods from six different categories: (1) Traditional augmentation methods, outputs from (2) CNN-based autoencoders, (3) super-resolution transformation, (4) frame interpolation, (5) video deblurring, and (6) colorization.
The dataset is split between a Two Alternative Forced Choice (2AFC) part and a Just Noticeable Differences (JND) part.
Examples from the two parts are shown in Fig.~\ref{fig:bapps_samples}.

% Example images from bapps

The 2AFC part consists of triplets of an image patch and two distorted versions of that patch.
Humans have judged each triplet according to which distorted version is more similar to the original.
The fraction of humans that preferred each version is used as the label for the triplet.
Part of the 2AFC dataset is designated for training, while the remainder is for evaluation.

The JND part consists of an image patch and a barely distorted version of that image.
Each pair were judged by humans that after a brief viewing had to decide if they were the same or not.
The pairs are labeled by the fraction of humans that thought they were the same.
The entire JND dataset is meant for evaluation.

%%%%%%%%%%%%%%%%%%%%%%%%%%%%%%%%%%%%%%%%%
% Methodology section
%%%%%%%%%%%%%%%%%%%%%%%%%%%%%%%%%%%%%%%%%

\section{Methodology}
\label{toc:methodology}

% Overview of all methodology

This work investigates whether the deep features of ImageNet~\cite{deng2009imagenet} pretrained CNNs contain the necessary information to adapt to different definitions of similarity and if this adaption can be achieved by learning scalars of the features for each definition.
To do this an altered version of the experiments by \textit{Zhang} \etal{}~\cite{zhang2018unreasonable} are used.

The experiments exhaustively test many combinations of loss networks, feature comparison methods, and training procedures.
Three loss networks pretrained on ImageNet~\cite{deng2009imagenet} with different architectures and feature extraction layers are used for extracting features.
Five methods for comparing the similarity of the extracted features are used to create metrics for each loss network.
The metrics are then either evaluated as baseline metrics without extra training or as adapted metrics trained on a specific ranking of distortions. 
Each combination is tested on 20 different random rankings of distortions.
With these combinations and each adapted metric being trained four times (to evaluate variance in training), this results in 1500 rows of collected data.
The combinations that have been evaluated are summarized in Table~\ref{tab:parameter_summary}.
All of these parts are detailed in the following subsection.

\begin{table}[t]
    \centering
    \caption{The different parameters evaluated in the experiments.}
    \label{tab:parameter_summary}
    \begin{tabular}{l l l l}
    \toprule
        Loss Networks & \makecell[l]{Comparison\\Methods} & Training & \makecell[l]{20 random\\rankings of:}\\\hline
        \makecell[lt]{AlexNet\\SqueezeNet~1.1\\VGG-16} &
        \makecell[lt]{Spatial\\Mean\\Sort\\Spatial+mean\\Spatial+sort} &
        \makecell[lt]{Pretrained\\baseline\\ \\Pretrained+\\adapted} &
        \makecell[lt]{Rotating\\Translating\\Lowering~brightness\\Shifting~hue\\Gaussian~blurring\\Zooming~in}\\
    \bottomrule
    \end{tabular}
    %\end{sc}
    %\end{footnotesize}
    %\end{center}
\end{table}

\subsection{Loss Networks}
% Loss networks and feature extraction
This work uses the same three loss networks as \textit{Zhang} \etal{}~\cite{zhang2018unreasonable}.
They are AlexNet~\cite{krizhevsky2014one}, SqueezeNet~1.1~\cite{iandola2016squeezenet}, and VGG-16~\cite{simonyan2015very} pretrained on the ImageNet~\cite{deng2009imagenet} dataset.
The specific implementation of each architecture and the trained model parameters were taken from the Torchvision~\cite{marcel2010torchvision} framework version $0.11.3$.
The features were extracted from layers throughout the convolutional parts of the models as detailed in Table~\ref{tab:networks_layers}.

\begin{table}[t]
    \centering
    \caption{Loss network architectures and feature extraction layers.}
    \label{tab:networks_layers}
    \begin{tabular}{l l}
    \toprule
        Architecture & Feature Extraction Layer\\\hline
        AlexNet~\cite{krizhevsky2014one} & \nth{1}, \nth{2}, \nth{3}, \nth{4}, and \nth{5} ReLU\\
        SqueezeNet 1.1~\cite{iandola2016squeezenet} & \nth{1} ReLU, \nth{2}, \nth{4}, \nth{5}, \nth{6}, \nth{7} and \nth{8} Fire\\
        VGG-16~\cite{simonyan2015very} & \nth{2}, \nth{4}, \nth{7}, \nth{10}, and \nth{13} ReLU\\
    \bottomrule
    \end{tabular}
    %\end{sc}
    %\end{footnotesize}
    %\end{center}
\end{table}

\subsection{Similarity Calculations}
% Comparison methods
The similarity between two images is calculated by using them each as input to the same loss network and then using the difference between the extracted deep features of each image as a distance metric.
There are many ways to compare the extracted deep features.
This work uses the methods used by \textit{Sjögren} \etal{}~\cite{sjogren2023identifying}, called spatial, mean, and sort comparisons.
The three comparison methods are detailed in Eq.~\ref{eq:spatial_comparison} to~\ref{eq:sort_comparison} below, where $z^l_x$ are the activations, which may have been channel-wise normalized, in layer $l$ from a loss network with input $x$ and extraction layers $l\in L$.
$\overline{z}$ and $z^\downarrow$ are the average and descending sorting of the channels in $z$ respectively.
$w_l$ are the scalars for the features of layer $l$, which are set to $1$ in the baseline cases and adapted to be positive values during adaption training, as explained later.

In addition to these three, the combined methods used by \textit{Sjögren} \etal{} are also used.
The combined methods consist of the sum of the spatial and mean metrics ($d_{spatial+mean} = d_{spatial} + d_{mean}$), as well as spatial and sort ($d_{spatial+sort} = d_{spatial} + d_{sort}$).

%Spatial comparisons (zhang2018unreasonable)
\begin{equation}
    \label{eq:spatial_comparison}
    d_{spatial}(x,x_0) = \sum_{l\in L}\frac{1}{C_lH_lW_l} ||w_l\odot(z^l_x-z^l_{x_0})||_2^2
\end{equation}

%Mean comparisons (kumar2022do)
\begin{equation}
    \label{eq:mean_comparison}
    d_{mean}(x, x_0) = \sum_{l\in L} \frac{1}{C_l} ||w_l\odot(\overline{z}^l_x - \overline{z}^l_{x_0})||^2_2
\end{equation}

%Sort comparisons (sjogren2023identifying)
\begin{equation}
    \label{eq:sort_comparison}
    d_{sort}(x, x_0) = \sum_{l\in L} \frac{1}{C_l} ||w_l\odot(z^{l\downarrow}_x - z^{l\downarrow}_{x_0})||^2_2
\end{equation}

\subsection{Distortions}
% The distortions and rankings used
To train and evaluate metrics for their ability to adapt to varying definitions of similarity, six distortions taken from commonly applied image augmentation procedures~\cite{jaiswal2021survey} are used.
The distortions are rotating, translating, lowering brightness, shifting hue, Gaussian blurring, and zooming in.
The distortions are implemented using the Torchvision~\cite{marcel2010torchvision} framework, and each time they are applied to an image, they do so with parameters chosen uniformly at random within given intervals.
The parameters and the intervals for their randomly chosen intervals for each distortion are shown in Table~\ref{tab:distortion_implementations}
Fig.~\ref{fig:svhn_and_stl10_distortions} shows random applications of the six distortion types to images from test sets of SVHN and STL-10.

\begin{table}[t]
    \centering
    \caption{Distortions and the intervals from which their parameters are randomly chosen.}
    \label{tab:distortion_implementations}
    \begin{tabular}{l l}
    \toprule
        Distortion & Parameters and intervals\\\hline
        Rotating & 30 to 330 degrees\\
        Translating & $-0.5$ to $0.5$ of the image size in each direction\\
        Lowering brightness & $0.1$ to $0.5$ of the original brightness\\
        Shifting hue & $-0.5$ to $0.5$ hue factor (all possible hues)\\
        Gaussian blurring & \makecell[l]{11 to 21 kernel size\\4 to 10 std. dev. for generating kernel values}\\
        Zooming in & 1.1 to 2 scale of zoom\\
    \bottomrule
    \end{tabular}
    %\end{sc}
    %\end{footnotesize}
    %\end{center}
\end{table}

\subsection{Adaption Training}
% Training procedure
Metrics are adapted to each ranking of the distortions by training the scalars ($w$) using the images from the SVHN training set.
For each image, two different distortions are chosen at random, and a triplet is created consisting of the original image and two distorted versions.
The triplet is labeled in the same way as a 2AFC triplet would be with 0 if the first distortion is earlier in the ranking and $1$ otherwise (\eg{} they are labeled according to which distortion should be considered more similar according to the ranking).
The metric being trained is then used to calculate the similarity scores between the original image and the two distortions.

During training, an auxiliary three-layer CNN is used in addition to the metric being trained.
Each layer consists of a $1\times1$ convolution with stride $1$ and leaky ReLU activation function, with the first two layers having 32 channels and the final only $1$.
The CNN takes five inputs; the similarity scores calculated by the metric between the original image and each of the distorted versions, the first score subtracted by the other, and each similarity score divided by the other.
The CNN gives an output between 0 and 1, according to which of the two is judged as more similar.
The CNN is trained along with the scalars because its judgment can be differentiated with respect to the similarity scores, which are needed for training.
The Binary Cross-Entropy (BCE) between the CNN output and the triplet label is used as the loss and backpropagated to update the CNN parameters as well as the scalars $w$ of the metric.
Training is performed for $10$ epochs with validation using 20\% of the training data.
During the last $5$ epochs, the learning rate decays linearly towards 0.
An additional synchronizing loss $\mathcal{L}_{sync}$ is used during training for each epoch until the validation 2AFC score is higher than random ($0.5$) to make sure that the scalars learn the correct similarity order instead of the opposite (both of which are equally useful to the CNN).
If the synchronization loss is removed, the adaption will likely learn the opposite similarity of some rankings since the CNN can invert that prediction when making its own. 
The loss is detailed in Eq.~\ref{eq:sync_loss} where $d$ is the metric being trained, $x$ is the original image, $x_0$ and $x_1$ are the distorted versions of $x$, $J$ is 1 if the distortion of $x_1$ is earlier in the ranking and 0 otherwise, and $\sigma$ is the sigmoid function.

%Sync loss
\begin{equation}
    \label{eq:sync_loss}
    \begin{split}
        \mathcal{L}_{sync}(x, x_0, x_1, J) = \\ 10\cdot\max(0,\text{BCE}(\sigma(d(x,x_0)-d(x,x_1)),J))
    \end{split}
\end{equation}

For each baseline metric and ranking four different adaptions are trained in order to measure the variance of training.

\subsection{Evaluation}
% How performance was measured and datasets
For each ranking and each metric, both baseline and adapted to that ranking, four performance scores were gathered.
The first two were gathered by taking the test set images in SVHN and STL-10 and creating 2AFC triplets consisting of the image and two versions of it distorted by two different randomly chosen distortions.
The 2AFC score of the metrics was calculated by whether they consider the version whose distortion is earlier in the ranking to be more similar.
The two remaining performance scores are the 2AFC and JND scores for the respective parts of the BAPPS dataset.
The calculation of the 2AFC score for a single sample is detailed in Eq.~\ref{eq:2afc score} for distance metric $d$, an image $x$, distorted versions $x_0$ and $x_1$, and the fraction $J$ of judgments that consider $x_1$ more similar to $x$ than $x_0$.
In the SVHN and STL-10 evaluations, $J$ is 0 if the distortion used for $x_0$ is earlier in the ranking and 1 otherwise.
The final 2AFC score is the average score for each sample.

%2afc score
\begin{equation}
    \label{eq:2afc score}
    \text{2AFC}(x, x_0, x_1, J) = 
    \begin{cases}
        J,& \text{if } d(x, x_1) < d(x, x_0)\\
        1-J,& \text{otherwise}
    \end{cases}
\end{equation}

%% file: sections/4_results.tex
%For tables see \footnote{\url{https://www.inf.ethz.ch/personal/markusp/teaching/guides/guide-tables.pdf} and \url{https://www.darkhorseanalytics.com/blog/clear-off-the-table apply religiously}}.
%As a general note, it is good practice to make a line break after each period in \LaTeX. 
%This makes it easier to read and modify, and especially to spot sentence construction flaws.

% Here is where you put evidence for all the claims you've made in the abstract - introduction - conclusion parts.
% All plots, tables and whatever else visualizations you have go here and gets to be discussed.
% Try to avoid marketing words such as VERY, EXTREMELY and so on.
% Do no hesitate to break down the section in multiple sub section to give some structure to the presentation of the results!

\section{Results and Analysis}
\label{toc:results}

\begin{figure}[t] % Always use [t] or [!t]
    \includegraphics[width=\columnwidth]{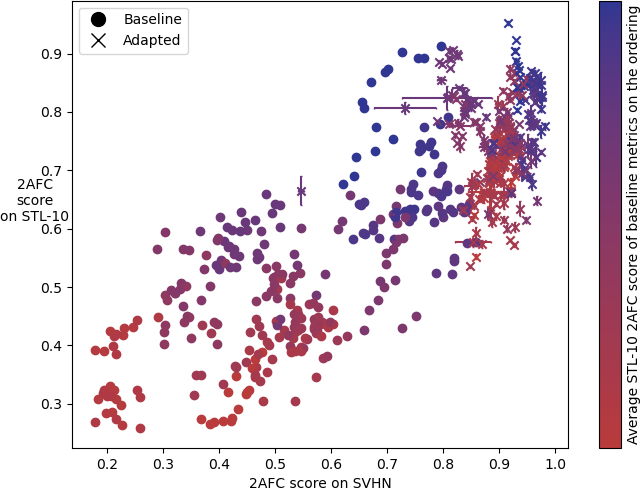}
    \caption{
       The 2AFC score of all metrics on SVHN and STL-10.
       The color of each point indicates which ranking the point was evaluated by going from the ranking with the lowest STL-10 2AFC score for baselines (red) to the highest (blue).
    }
    \label{fig:svhn_vs_stl10_for_all_points}
\end{figure}

The performance scores for correctly ranking the distortions on SVHN and STL-10 are visualized in Fig.~\ref{fig:svhn_vs_stl10_for_all_points}.
The figure shows the score for each combination of ranking, loss network, comparison method, and whether the metric has been adapted ($\times$) or not ($\bullet$).
The adapted metrics are shown as the average and standard deviation (often close to 0) of the four trained metrics.
The data points are colored according to what ranking they were evaluated on, with the ranking that baseline models on average had the lowest STL-10 2AFC score being red, the highest score being blue, and a gradient for the rankings in between.

% Scatterplot of svhn vs stl-10 for all datapoints colored by order

It is not so surprising that metrics that are better at aligning with a specific ranking for SVHN images also tend to do better for that ranking on STL-10 images.
However, the two datasets contain quite different images, especially when it comes to coloration.
This might be why almost all adapted metrics outperform the best baseline metric on SVHN, while the worst adapted metrics perform about the same as the average baseline on STL-10.
Still out of 300 adapted metrics only 2 do not outperform their baseline counterparts on average, and in both cases the performance difference is $\sim0.01$.

The adapted metrics also tend to perform worse and better on the same rankings as the baseline metrics do, with the Spearman correlation between the baseline and adapted average performances for each ranking being $0.66$ and $0.72$ for SVHN and STL-10 respectively.
This would likely not have been the case if the adapted metrics had been allowed to learn negative scalars, since inverting the scalars produces a metric that with the same performance if the ranking is reversed.
In short, a metric that performs worse than random can be improved by simply inverting its predictions.
In fact, the worst adapted metrics are only barely better than random chance on STL-10 and would be significantly outperformed by the inversions of their baselines.
These poorly adapted metrics consist almost exclusively of VGG-16 architectures and have a significantly better performance on SVHN.
This suggests that the poor performance is not due to lacking the features needed to learn the rankings but instead seems to be a case of poor generalization to another image dataset with the same ranking.
In general, this implies that the most difficult rankings to learn would actually be those where the baseline models achieve close to random performance since inversions would not significantly improve performance.

\begin{figure*}[t] % Always use [t] or [!t]
    \begin{subfigure}[b]{\columnwidth}
        \centering
        \includegraphics[width=\columnwidth]{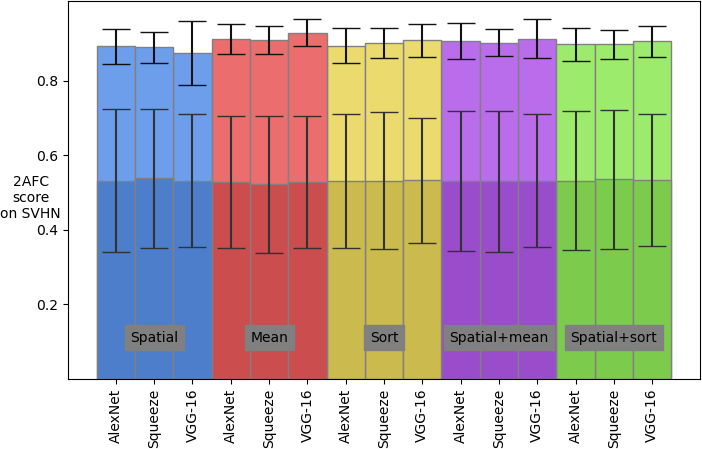}
        \caption{}
    \end{subfigure}
    \begin{subfigure}[b]{\columnwidth}
        \centering
        \includegraphics[width=\columnwidth]{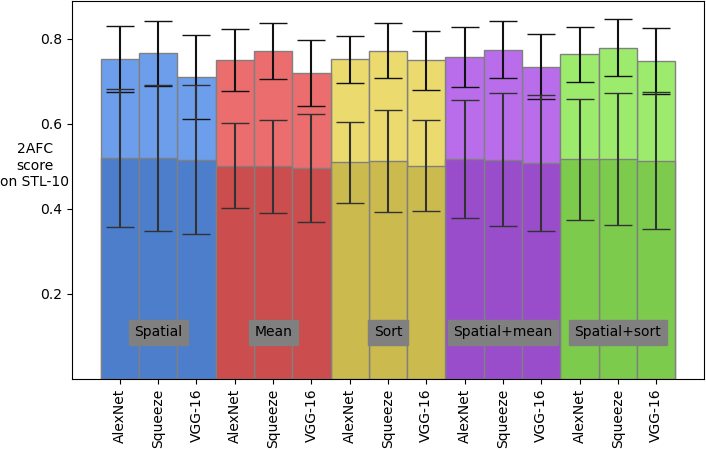}
        \caption{}
    \end{subfigure}
    \caption{
       The 2AFC score on (a) SVHN and (b) STL-10 for each loss network, comparison method, and whether or not it was adapted, averaged over all rankings of the six distortions.
       The lower bars are the baseline models and the upper their adapted counterparts.
    }
    \label{fig:score_avg_orders}
\end{figure*}

Fig.~\ref{fig:score_avg_orders} illustrates the performance on SVHN and STL-10 images of the different loss networks and comparison methods for baseline (lower bars) and adapted metrics (upper bars).
The performance is shown as the average and standard deviation over all rankings.
The figure indicates that the specific loss network and comparison method does not significantly impact results, at least among the networks and methods evaluated in this work.

% Barplot of each metric types average ranking performance

It is clear that the adapting metrics provide a significant advantage on the average ranking.
Going from close to random performance ($0.5$) which is expected from the baselines to $0.90$ and $0.75$ on average for the adapted metrics on SVHN and STL-10.
The adapted metrics also have lower standard deviations as they all perform above random and therefore are less spread out among the possible scores.
Of the 300 combinations of loss networks, comparison methods, and rankings the adapted metrics are outperformed by their baseline counterparts two times on STL-10 and never on SVHN.

%VGG generalizes poorly, why?
The VGG-16 metrics have a slight advantage on SVHN, that advantage does not generalize to STL-10.
This could potentially be because the features extracted from the VGG-16 network are easier to overfit the scalars on.
In general, overfitting occurs more often for larger models.
While spatial and sort DPS on VGG-16 has roughly twice as many features as SqueenzeNet and four times as many features as Alexnet, this is not the case for mean DPS where SqueezeNet has by far the most features.

%Comparison method does not matter so much despite distortions they are known to be impacted by, why?
Another noteworthy detail is that the choice of comparison method does not seem to have a significant impact on performance.
This is surprising since previous work has shown that spatial DPS metrics struggle with translation and rotation~\cite{sjogren2023identifying}, both of which are included as distortions in the rankings.
A potential answer could have been that the different comparison methods perform well on different rankings, with spatial performing better when translation is important and worse otherwise.
However, this is not the case as there is a strong correlation between the performance on different rankings of all comparison methods.
The Spearman correlation is above $0.9$ between all comparison methods' STL-10 performances on the rankings.
Another potential answer is that the metrics have adapted to some quirks of the distortions that are not affected as heavily by the issues with translation and rotation.
For example, the default behavior of translation and rotation in Torchvision which was used in this work colors the missing pixels black.
The metrics might have adapted to discover if the edges or corners of the image are black and then weigh that according to where the two distortions show up in the ranking.
Such a quirk would be discoverable by all comparison methods, making their differences less impactful.

%\todo{Consider writing about which ranking/distortions are difficult, which are close to random, and which are easy?}

%About the effect on 2AFC and JND performance
The adaption training is slightly detrimental to the metrics' performance on both the 2AFC and the JND parts of the BAPPS dataset.
Both scores lower by $\sim0.01$ on average for the adapted metrics compared to their baseline counterparts.
It is interesting that the average baseline outperforms the adapted metrics on BAPPS on all rankings.
It would seem that certain rankings would align better with the average perception of the subjects that gathered judgments for the BAPPS dataset, and therefore that adapting to those rankings would improve performance.
However, the results show that the adaption procedure used in this work alter some of the features that were useful for BAPPS too much.
This suggests that fine-tuning the entire loss network might be even more detrimental than the simple scalars learned in this work as it would have the potential to completely alter the features used.

To test whether fine-tuning the loss network would be even more detrimental to the BAPPS scores an additional smaller run of experiments was conducted where DPS metrics with AlexNet architecture and spatial comparison were adapted by fine-tuning the parameters of the AlexNet model, in addition to training the scalars.
The results from this trial showed a significant improvement in the 2AFC score on the SVHN and STL-10 images but with additional detriment to the performance on BAPPS.
On average, compared to only using scalars, the SVHN and STL-10 scores improved by $0.10$ and $0.05$, while the 2AFC and JND scores on BAPPS decreased by $0.02$ and $0.01$.
A significant part of the performance boost on SVHN and STL-10 can be attributed to fine-tuning, allowing the inversion of features that the positive scalars could not perform.
This is made clear by the lower correlation between which rankings the baselines perform well at compared to which the scalars-only and fine-tuned metrics perform well on.
The Spearman correlation for the baseline and scalars-only is $0.70$ and it is $0.35$ for the baseline and fine-tuning.

% Since the metrics seem to perform roughly the same irregardless of architecture and comparison methods, maybe only show results from spatial after the initial presentation?

%% file: sections/5_analysis.tex
\section{Discussion}
\label{toc:discussion}
% Here we want to write about some important aspects of your observations. These are differing, depending on the paper, e.g.,
% Some typical good cases and failure cases; or cases where the methods disagree
% Looking into the actual activations of several layers, what is important for the method to make it work. Is there any unwanted bias towards the data
% Research is not about performance, but about hypotheses evaluation, investigation, and understanding. 
% What does the results and analysis lead us to believe, but that we cannot yet prove.
% How do we think the field needs to develop?
% Other interesting things that can be discussed that aren't necessarily analysis of the results

% Start with the implications of the results
%This work is proof-of-concept in nature, showing what is possible and giving a starting point for further exploration.
%\todo{Maybe discuss more about the point of the work?}

The results show that it is possible for DPS metrics to be adapted through learning positive scalars for the extracted features to a definition of similarity given by which distortions should be perceived as more similar.
This suggests that, in general, the features of ImageNet pretrained CNNs contain the information needed to adapt to different contexts simply by weighing them differently.
The learned adaptions also generalize to images from a significantly different dataset.
Though, the performance is not great for all rankings, especially on the images from STL-10, on which the metrics were not adapted.
Some of the lackluster performance can be attributed to forcing positive scalars.
Even with negative scalars allowed, the rankings on which the baseline metrics have close to random performance would likely still be difficult.
However, the adapted metrics still perform significantly better than random in these cases.

Learning the adaptions is also shown not to be significantly detrimental to the performance on BAPPS.
This is desirable because it allows adapting metrics to specific contexts without significant risk of losing other desirable properties.
For example, a metric could be adapted to deal with some specific invariances in the given data on a very specific training set without having to simultaneously train it on the original data to keep performance from collapsing.
This can otherwise be an issue in settings where models are continuously updated~\cite{french1999catastrophic}.
When metrics were adapted by also fine-tuning the loss network, this was further detrimental to performance.
Perhaps then, a better approach to adaptability is to integrate the adaption with the pretraining of the loss network.
\textit{Kumar} \etal{}~\cite{kumar2022do} have shown that the pretraining procedure has a significant impact on perceptual similarity performance.
Including the adaption context among the other pretraining data or even specifically pretraining on a dataset specific to the domain could improve performance on the adapted context without degrading performance on general perceptual similarity datasets.

It is also noted that the metrics might not be adapting to measure the similarity of images but rather to distinguish different distortion types from each other.
If this is the case, it likely arises from only training on triplets from the same image, meaning that being able to classify distortions is enough to achieve high accuracy.
This could be solved by taking inspiration from contrastive learning, which forms negative pairs from different images.
For the training presented here, this could take the form of additionally including triplets where one of the images is the distorted version of other images.
The distorted version of the same image would then be considered more similar regardless of which distortion is applied to the other.
Interestingly, this issue would also be present in BAPPS training that has been conducted by prior works~\cite{zhang2018unreasonable, kettunen2019robust, kumar2022do}.
The same solution would be applicable in this case as well, which could likely improve performance on the dataset even further.

\section{Future Work}
\label{toc:future}
%% Future Work

There are many ways to build on this work as well as the issues posed by ambiguity in similarity.
The adaption method explored in this work could be applied to a more realistic scenario instead of the proof-of-concept scenario that was used here.
The images from of medical histopathology and non-RGB sensors where the perception of similarity from natural images might not neatly apply are interesting test cases.
For these domains and other applications, the question of how to get the data for adaption training is raised.
Perhaps a similar use of distortions is applicable, where the order of similarity is defined by experts in the field.

Similar questions are being explored in the field of contrastive learning, where feature extraction models are trained to learn similar features for similar data, which is similar to what is done when DPS metrics are trained.
In contrastive learning, distortions are used to learn that an object is the same even as the image of it is distorted.
However, in that field, it has been noted that the distortions that work well for natural images from ImageNet, do not generalize to other domains such as medical imaging~\cite{chandra2023self}.
Instead, using the information in the data to determine which images to consider similar is being used~\cite{chandra2023self}.
For perceptual similarity, it is not desirable to learn that two distortions of an image are the same, but similar approaches to finding how similar two images should be considered might be applicable.

DPL is another domain where adapting the metric for a given definition of similarity might be useful.
Loss functions inherently have a usefulness associated with how well they work to train models for a given task and adapting DPL to that task might be beneficial.
Loss functions also tend to work in multiple different settings throughout training.
Early the output will likely be very dissimilar to the desired output, while towards the end of the training, the two are hopefully very similar.
What is more important for achieving similarity and performance in these different settings might be very different.

The question of ambiguity might not be as interesting in cases where images only have barely noticeable differences.
If humans struggle to even notice the differences then finding out which slight difference is less similar might not be a considerable issue.
Additionally, in the domain of very similar images older similarity metrics that rely on pixels likely work very well as the differences of the images are likely on the pixel-level rather than the more noticeable structural level.

Conversely, tasks where images with more significant and noticeable differences are likely better suited for exploring the impact of ambiguity and adaptability.
For example, when performing an image reverse search the desired result can vary heavily even for the same input image.
One user might want to find another version of the same image while another wants to find images with the same composition, and a third wants to find scenes containing similar objects.
In these scenarios having metrics that give scores based on different contexts would be useful.
Additionally, the field of image retrieval has already acknowledged the issues that come with ambiguity~\cite{saha2004agent, rossetto2016dealing}.
The task of image retrieval would therefore be an interesting case to further study.